\pdfoutput=1
\documentclass[11pt,a4paper]{article}
\usepackage[hyperref]{emnlp2020}
\usepackage{times}
\usepackage{latexsym}

\usepackage{booktabs}
\usepackage{graphicx}
\usepackage{lipsum}
\usepackage{amsmath}
\usepackage{amssymb}
\usepackage{textcomp}
\usepackage{breakurl}
\usepackage[
  separate-uncertainty = true,
  multi-part-units = repeat
]{siunitx}

\usepackage{microtype}

\aclfinalcopy 
\def\aclpaperid{1146} 

\newcommand{\Ni}{({\em i})~}
\newcommand{\Nii}{({\em ii})~}
\newcommand{\Niii}{({\em iii})~}

\newcommand{\en}{$en$}
\newcommand{\fr}{$fr$}
\newcommand{\es}{$es$}
\newcommand{\de}{$de$}

\newcommand{\ssnmt}{SSNMT}

\title{Self-Induced Curriculum Learning \\ in Self-Supervised Neural Machine Translation}

\author{Dana Ruiter \\
Saarland University \\
DFKI GmbH \\ 
\And
   Josef van Genabith \\
   Saarland University \\
   DFKI GmbH \\
   \texttt{druiter@lsv.uni-saarland.de} \\
   \texttt{\{josef.van\_genabith,cristinae\}@dfki.de} \And
Cristina Espa{\~n}a-Bonet \\
   DFKI GmbH \\
}

\date{}

\begin{document}
\maketitle
\begin{abstract}
Self-supervised neural machine translation (\ssnmt) jointly learns to identify and select suitable training data from comparable (rather than
parallel) corpora and to translate, in a way that the two tasks
support each other in a virtuous circle. In this study, we
provide an in-depth analysis of the sampling choices the \ssnmt\ model
makes during training. We show how, without it having been told to do
so, the model self-selects samples of increasing \Ni complexity and
\Nii task-relevance in combination with \Niii performing a denoising curriculum. We observe that the dynamics of the mutual-supervision signals of both system internal representation types are vital for the extraction and translation performance. 
We show that in terms of the Gunning-Fog Readability index, \ssnmt\ starts extracting and learning from Wikipedia data suitable for high school students and quickly moves towards content suitable for first year undergraduate students.
\end{abstract}

\section{Introduction}
\label{s:intro}

Human learners, when faced with a new task, generally focus on simple examples before applying what they learned to more complex instances. This approach to learning based on sampling from a curriculum of increasing complexity has also been shown to be beneficial for machines and is referred to as \emph{curriculum learning} (CL) \citep{bengio_curricullum-learning_2009}. 
Previous research on curriculum learning has focused on selecting the best distribution of data, i.e. order, difficulty and closeness to the final task, to train a system. In such a setting, data is externally prepared for the system to ease the learning task. In our work, we follow a complementary approach: we design a system that selects by itself the data to train on, and we analyse the selected distribution of data, order, difficulty and closeness to the final task, without imposing it beforehand.
Our method resembles \emph{self-paced learning} (SPL) \citep{kumar2010self-paced}, in that it uses the emerging model hypothesis to select samples online that fit into its space as opposed to most curriculum learning approaches that rely on judgements by the target hypothesis, i.e. an external \emph{teacher} \citep{hacohen2019power} to design the curriculum.

We focus on machine translation (MT), in particular, self-supervised machine translation (\ssnmt) \citep{ruiter-etal-2019-self}, which exploits the internal representations of an emergent neural machine translation (NMT) system to select useful data for training, where each selection decision is dependent on the current state of the model. Self-supervised learning \citep{rainaEtAl:2007,bengioEtAl:2013} involves a primary task, for which labelled data is not available, and an auxiliary task that enables the primary task to be learned by exploiting supervisory signals within the data. In \ssnmt, both tasks, data extraction and learning MT, enable and enhance each other. This and the mutual supervision of the two system internal representations lead to a self-induced curriculum, which is the subject of our investigation. 

In Section~\ref{s:sota} we describe related work on CL, focusing on MT. Section~\ref{s:ssnmt} introduces the main aspects of
self-supervised neural machine translation. 
Here, we analyse the performance of both the primary and the auxiliary tasks. 
This is followed by a detailed study of the self-induced curriculum in Section~\ref{s:cl} where we analyse the characteristics of the distribution of training data obtained in the auxiliary task of the system. We conclude and present ideas for further work in Section~\ref{s:conclusion}.

\section{Related Work}
\label{s:sota}

\textbf{Machine translation} has experienced major improvements in translation quality due to the introduction of 
neural architectures~\citep{cho2014learning,bahdanau2014neural,vaswani2017attention}.
However, these rely on the availability of large amounts of parallel data. To overcome the need for labelled data, unsupervised neural machine translation (USNMT) \citep{lampleEtAl:ICLR:2018,artetxeEtAl:ICLR:2018,yangEtAl:2018} focuses on the exploitation of very large amounts of monolingual sentences by combining denoising autoencoders with back-translation and multilingual encoders. Further combining these with phrase tables from statistical machine translation leads to impressive results \citep{lampleEtAl:EMNLP:2018, artetxeEtAl:EMNLP:2018, ren2019unsupervised, artetxe2019effective}.  USNMT can be combined with pre-trained language models (LMs) \citep{lample_cross-lingual_2019, song2019mass, liu2020multilingual}. \citet{brown2020language} train a very large LM on billions of monolingual sentences which allows them to perform NMT in a few-shot setting.
Self-supervised NMT (\ssnmt) \citep{ruiter-etal-2019-self} is an alternative approach focusing on \emph{comparable}, rather than \emph{parallel} data.
The internal representations of an emergent NMT system are used to identify useful sentence pairs in comparable documents. Selection depends on the current state of the model, resembling a type of self-paced learning \citep{kumar2010self-paced}.

Data selection in \ssnmt\ is directly related to \textbf{curriculum learning}, the idea of presenting training samples in a \emph{meaningful} order to benefit learning, e.g. in the form of faster convergence or improved performance \citep{bengio_curricullum-learning_2009}. Inspired by human learners, \citet{elman1993learning} argues that a neural network's optimization can be accelerated by providing samples in order of increasing complexity. 
While \textbf{sample difficulty} is an intuitive measure on which to base a learning schedule, curricula may focus on other metrics such as \textbf{task-relevance} or \textbf{noise}.

To date, \textbf{curriculum learning in NMT} has had a strong focus on the relevance of training samples to a given translation task, e.g. in domain adaptation.
\citet{vanderWees2017dynamic} train on increasingly relevant samples while gradually excluding irrelevant ones.
They observed an increase in BLEU over a static NMT baseline and a significant speed-up in training as the data size is incrementally reduced. \citet{zhang2019curriculum} adapt an NMT model to a domain by introducing increasingly domain-distant (\emph{difficult}) samples. This seemingly contradictory behavior of benefiting from both increasingly difficult (domain-distant) \emph{and} easy (domain-relevant) samples has been analyzed by \citet{weinshall2018curriculum}, showing that the initial phases of training benefit from easy samples with respect to a hypothetical competent model (\emph{target hypothesis}), while also being \emph{boosted} \citep{freund1996boosting} by samples that are difficult with respect to the current state of the model \citep{hacohen2019power}. 
In \citet{wang2019dynamically}, both domain-relevance and denoising are combined into a single curriculum.

The denoising curriculum for NMT proposed by \citet{wang2018denoising} is related to our approach in that they also use \emph{online data selection} to build the curriculum based on the current state of the model. However, the noise scores for the dataset at each training step depend on fine-tuning the model on a small selection of clean data, which comes with a high computational cost. To alleviate this cost, \citet{kumar2019reinforcement} use reinforcement learning on the pre-scored noisy corpus to jointly learn the denoising curriculum with NMT. In Section~\ref{s:controlled} we show that our model exploits its self-supervised nature to perform denoising by selecting parallel pairs with increasing accuracy, without the need of additional noise metrics.

Difficulty-based curricula for NMT that take into account sentence length and vocabulary frequency have been shown to improve translation quality when samples are presented in increasing complexity \citep{kocmi2017curriculum}. \citet{platanios2019competence} link the introduction of difficult samples with the NMT models' \emph{competence}. Other difficulty-orderings have been explored extensively in \citet{zhang2018empirical}, showing that they, too, can speed-up training without a loss in translation performance.

\ssnmt\ jointly learns to find and extract similar sentence pairs from comparable data and to translate.
The extractions can be compared to those obtained by \textbf{parallel data mining} systems where strictly parallel sentences are expected. 
Beating early feature-based approaches, sentence representations obtained from NMT systems or tailored architectures are achieving a new state-of-the-art in parallel sentence extraction and filtering
\citep{espana2017empirical,gregoire-langlais-2018-extracting,artetxe2018margin, hangyaFraser:2019,chaudharyEtal:2019}. Using a highly multilingual sentence encoder, \citet{schwenk2019wikimatrix} scored Wikipedia sentence pairs across various language combinations (\emph{WikiMatrix}). 
Due to its multilingual aspect and the close similarity with the raw Wikipedia data we use, we also use scored WikiMatrix data for one of the comparisons (Section~\ref{s:controlled}).

\begin{table*}[t]
\small
\centering
\setlength{\tabcolsep}{2pt}
\begin{tabular}{l @{\hspace{1em}} rrc @{\hspace{1em}} rrc   @{\hspace{1em}} rr @{\hspace{1em}} rr} 
\toprule
       & \multicolumn{3}{c}{WP, ~L1} & \multicolumn{3}{c}{WP, ~L2} & \multicolumn{2}{c}{EP, ~L1} & \multicolumn{2}{c}{EP, ~L2} \\
       \cmidrule(r){2-4} \cmidrule(r){5-7}  \cmidrule(r){8-9}  \cmidrule(r){10-11}
L1--L2       & \# Sent.  & \# Tokens  & Sent./Article & \# Sent.  & \# Tokens  & Sent./Article  & \# Sent.  & \# Tokens  & \# Sent.  & \# Tokens \\ 
\midrule
\en--\fr  & 117 / 42    & 2693/1205 & 28 & 38/25 &  644/710 & 16 & 1+6 & 25+80  & 1+3 & 27+87  \\
\en--\de  & 117 / 37    & 2693/987 & 29  & 51/30 & 1081/742 & 24 & 1+9 & 25+180 & 1+7 & 26+192  \\
\en--\es  & 117 / 35    & 2693/937 & 32  & 27/20 &  691/572 & 17 & 1+7 & 24+84  & 1+4 & 26+91\\
\bottomrule
\end{tabular}
\caption{Millions of sentences and tokens for the corpora used. For Wikipedia (WP), we report the sizes for both the monolingual/comparable editions; for Europarl (EP), true+false splits (see Section~\ref{s:controlled}). 
}
\label{t:corpora}
\end{table*}

\section{Self-Supervised Neural Machine Translation (\ssnmt)}
\label{s:ssnmt}

\ssnmt\ is a joint data selection and training framework for machine translation, introduced in \citet{ruiter-etal-2019-self}. \ssnmt\ enables learning NMT from \emph{comparable} rather than parallel data, where comparable data is a collection of multilingual topic-aligned documents.\footnote{Wikipedia is an example; the French article on \emph{Paris} is different from the German one. They are not translations of each other, but they are on the same topic.} Its basic architecture uses the semantic information encoded in the internal representations of a standard NMT system to determine at training time if an input sentence pair is \emph{similar enough} or not, and therefore whether it should be used for training or not. Selection is made online, so, the more the semantic representations improve during training, the more truly parallel sentence pairs are selected. Because of this, the nature of the selected pairs naturally evolves during training, and this evolution is what we analyze as self-induced curriculum learning in Section~\ref{s:cl}. 

\ssnmt\ is based on a bidirectional NMT system $\{L1, L2\} \rightarrow \{L2, L1\}$ where the engine learns to translate simultaneously from a language $L1$ into another language $L2$ and vice-versa with a single encoder and a single decoder. This is important in the self-supervised architecture because it represents the two languages in the same semantic space. In principle, the input data to train the system is a monolingual corpus of sentences in $L1$ and a monolingual corpus of sentences in $L2$ and the system learns to find and select similar sentence pairs. In order to speed-up training, we use a comparable corpus such as Wikipedia, where we can safely assume that there are comparable (similar) and parallel sentence pairs in related documents ${D_{L1}, D_{L2}}$.

Given a document pair ${D_{L1}, D_{L2}}$, the \ssnmt\ system encodes each sentence of each document into two fixed-length vectors $C_w$ and $C_h$
\begin{equation}
    C_{w} = \sum_{t} w_t, ~~~~~~~~~
    C_{h} = \sum_{t} h_t,
\end{equation}
\noindent where $w_t$ is the word embedding and $h_t$ the encoder output at time step $t$. For each of the \emph{sentence representations} $s$, all combinations of sentences $s_{L1} \times s_{L2} \| s_{L1} \in D_{L1}$ and $s_{L2} \in D_{L2}$ are encoded and scored using the \emph{margin-based} measure by \citet{artetxe2018margin} with $k=4$.

What follows is a selection process, that identifies the top scoring $s_{L2}$ for each $s_{L1}$ and vice-versa. If a pair $\{s_{L1}, s_{L2}\}$ is top scoring for both language directions \emph{and} for both sentence representations, it is accepted without involving any hyperparameter or threshold. This is the high precision, medium recall approach in \citet{ruiter-etal-2019-self}. Whenever enough pairs have been collected to create a batch, the system trains on it, updating its weights, improving both its translation and extraction ability to fill the next batch.

\begin{table*}[t]
\centering
\small
\begin{tabular}{l @{\hspace{2em}} ccc ccc @{\hspace{2em}} cc}
\toprule
    & \multicolumn{6}{c}{\ssnmt} & \multicolumn{2}{c}{SotA} \\
   \cmidrule(r){2-7}\cmidrule(r){8-9}
    & \multicolumn{3}{c}{L1-to-L2} & \multicolumn{3}{c}{L2-to-L1} & L1-to-L2 & L2-to-L1\\
     \cmidrule(r){2-4}\cmidrule(r){5-7}\cmidrule(r){8-8}\cmidrule(r){9-9}
 L1--L2  & BLEU   & TER    & METEOR  & BLEU   & TER    & METEOR  & BLEU & BLEU \\ \midrule
  \en--\fr  & 29.5$\pm$.6  & 51.9$\pm$.6  & 46.4$\pm$.6   & 27.7$\pm$.6   & 53.4$\pm$.7  & 30.3$\pm$.4   & 45.6/25.1/37.5 & --/24.2/34.9 \\ 
 \en--\de   & 15.2$\pm$.5  & 68.5$\pm$.7  & 30.3$\pm$.5   & 21.2$\pm$.6  & 62.8$\pm$.9  & 25.4$\pm$.4   & 37.9/17.2/28.3 & --/21.0/35.2 \\ 
 \en--\es   & 28.6$\pm$.7  & 52.6$\pm$.7  & 47.8$\pm$.7   & 28.4$\pm$.7  & 54.1$\pm$.7  & 30.5$\pm$.4  & --/--/-- & --/--/-- \\ 
\bottomrule
\end{tabular}
\caption{Automatic evaluation of \ssnmt\ on NT14 (\fr) NT16 (\de) NT13 (\es).
Most right columns show the comparison with three SotA systems for supervised NMT \citep{edunov2018understanding} / USNMT \citep{lampleEtAl:EMNLP:2018} / pre-trained+LM USNMT \citep{song2019mass}.
}
\label{t:results_wiki}
\end{table*}

\subsection{Translation Quality}
\label{ss:tq}

\paragraph{Experimental Setup}
\label{s:data}
We use Wikipedia (WP) as a comparable corpus and download the English, French, German and Spanish dumps,%
\footnote{Dumps were downloaded on January 2019 from \url{dumps.wikimedia.org/}} pre-process them and extract comparable articles per language pair using \texttt{WikiTailor}\footnote{\url{github.com/cristinae/WikiTailor}}
\citep{barron2015factory,espanaEtAl:2020}. 
All articles are normalized, tokenized and truecased using standard \texttt{Moses} \citep{koehn2007} scripts. For each language pair, a shared byte-pair encoding (BPE) \citep{sennrich2015neural} of 100\,$k$ merge operations is applied. Following \citet{johnson-etal-2017-googles}, a language tag is added to the beginning of each sequence. 

The number of sentences, tokens and average article length is reported in Table~\ref{t:corpora}.
For validation we use \emph{newstest2012} (NT12) and for testing \emph{newstest2013} (NT13) for $en$--$es$ and \emph{newstest2014} (NT14) or \emph{newstest2016} (NT16) for $en$--$\{fr,de\}$.
The \ssnmt\ implementation\footnote{\url{github.com/ruitedk6/comparableNMT}} builds on the transformer base \citep{vaswani2017attention} in \texttt{OpenNMT} \citep{opennmt}. 
All systems 
are trained
using a batch size of 50 sentences with maximum length of 50 tokens.

Monolingual embeddings trained using \texttt{word2vec} \citep{mikolov2013distributed}\footnote{\url{github.com/tmikolov/word2vec}} 
on the complete WP editions are projected into a common multilingual space via \texttt{vecmap}\footnote{\url{github.com/artetxem/vecmap}}
\citep{artetxe2017acl} to attain bilingual embeddings between $en$--\{$fr$,$de$,$es$\}. These initialise the NMT word embeddings ($C_w$).

As a control experiment and purely in order to analyse the quality of the \ssnmt\ data selection auxiliary task, we use the Europarl (EP) corpus \citep{koehn2005europarl}. The corpus is pre-processed in the same way as WP, and we create a synthetic comparable corpus from it as explained in Section \ref{s:controlled}. For these experiments, we use the same data for validation and testing as mentioned above.

\paragraph{Automatic Evaluation}
\label{ss:evalGlobal}

We use BLEU \citep{papineni2002BLEU}, TER \citep{snover2006astudy} and METEOR \citep{lavie2007meteor} to evaluate translation quality. 
For calculating BLEU, we use \texttt{multi-bleu.perl}, while TER and METEOR are calculated using the \texttt{scoring} package\footnote{\url{kheafield.com/code/scoring.tar.gz}} which also provides confidence scores.
\ssnmt\ translation performance training on the $en$--$\{fr,de,es\}$ comparable Wikipedia data is reported in Table~\ref{t:results_wiki} together with a comparison to the current state-of-the-art (SotA) in supervised and (pre-trained) USNMT. \ssnmt\ is on par with the current SotA in USNMT, outperforming it by 3--4 BLEU points in \en--\fr\ with lower performance on \en--\de\ ($\sim$3 BLEU). Note that unsupervised systems such as  \citet{lampleEtAl:EMNLP:2018} use more than $400\,M$ monolingual sentences for training while \ssnmt\ uses an order of magnitude less by exploiting comparable corpora. However, once unsupervised NMT is combined with LM pre-training, it outperforms \ssnmt\ (which does not use LM pre-training) by large margins, i.e. around 7 BLEU points for \en--\fr\ and 13 BLEU for \en--\de.

\subsection{Data Extraction Quality}
\label{s:controlled}

\paragraph{Experimental Setup}
To get an idea of the data extraction performance of an \ssnmt\ system, we perform control experiments on synthetic comparable corpora, as there is no underlying ground truth to Wikipedia. For these purposes, we use the $en$--\{$fr$,$de$,$es$\} versions of Europarl. After setting aside $1M$ parallel pairs as \emph{true} samples to evaluate \ssnmt\ data extraction performance, the target sides of all remaining source-target pairs in EP are scrambled to create non-parallel (\emph{false}) source-target pairs.
In order to keep the synthetic comparable corpora close to the statistics of the original comparable Wikipedias, we control the EP true:false (parallel:non-parallel) sentence pair ratio to mimic the ratios we observe in our extractions from WP.
We assume that all WP sentences accepted by \ssnmt\ are true (parallel) examples, and that the number of false examples (non-parallel) are the rejected ones.
With this, we estimate base true:false ratios of 1:4 for $en$--\{$fr$,$es$\} and 1:8 for $en$--$de$.%
\footnote{In a manual evaluation annotating 10 randomly sampled WP articles for L1 and L2 in $en$--\{$fr$,$es$,$de$\} each, the true:false ratios resulted 3:8 for $en$--$fr$, 1:4 for $en$--$es$ and 1:8 for $en$--$de$ which validate the assumption.}
The false samples created from EP are oversampled in order to meet this ratio given that there are $1M$ true samples. Further, we calculate the average article length of the comparable WPs and split the synthetic comparable
samples into pseudo-articles with this length. 
The statistics of the synthetic pseudo-comparable EPs are reported in Table~\ref{t:corpora}.
We then train and evaluate the \ssnmt\ system on the synthetic comparable data. 

\begin{figure}
    \centering
    \includegraphics[width=1.0\linewidth]{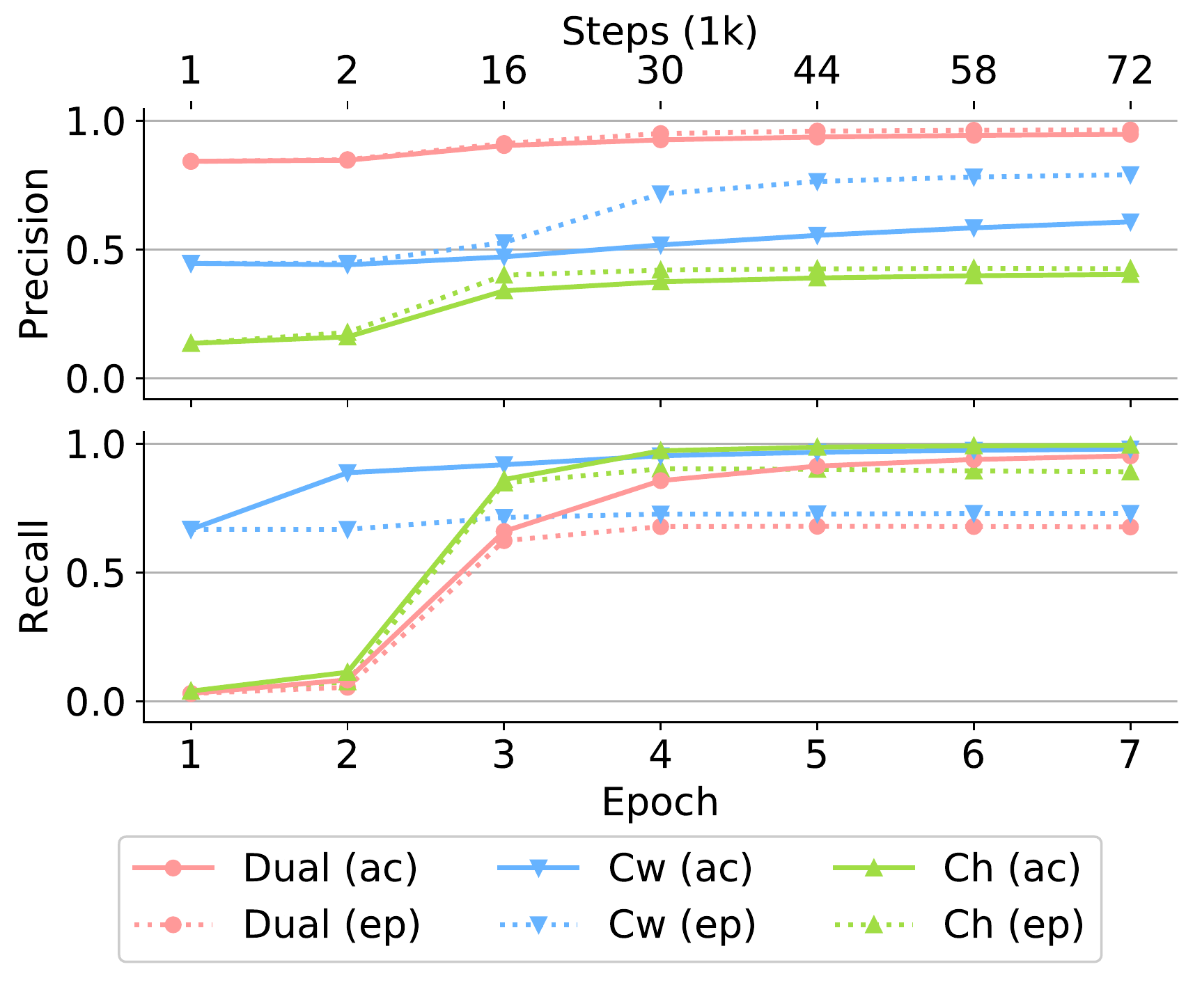}
    \caption{Accumulated (ac) and epoch-wise (ep) precision and recall on the $en$--$fr$ EP-based synthetic comparable data. 
    }
    \label{f:control}
\end{figure}

\paragraph{Automatic Evaluation}
The pairs \ssnmt\ extracts from the pseudo-comparable EP articles at each epoch are compared to the 1$M$ ground truth pairs to calculate \emph{epoch-wise} extraction precision (P) and recall (R). Further, we also take the concatenation of all extracted sentences from the very beginning up to a certain epoch in training in order to report \emph{accumulated} P and R. As we are interested in the final extraction decision based on the intersection of both representations $C_w$ and $C_h$ (\emph{dual}), but also in the decisions of each single representation ($C_w$, $C_h$), we report the performance for all three representation combinations on EP$_{enfr}$ in Figure~\ref{f:control}. Similar curves are observed for EP$_{ende}$ and EP$_{enes}$, which are considered in the discussion below.

At the beginning of training, the extraction \textbf{precision} of each representation itself is fairly low with P$\in$[0.45,0.66] for $C_w$ and P$\in$[0.14,0.40] for $C_h$. The fact that $C_w$ is initialized using pre-train-ed embeddings, while $C_h$ is not, leads to the large difference in initial precision between the two. As both representations are combined via their intersections, the final decision of the model is high precision already at the beginning of training with values between $0.78$--$0.87$. As training progresses and the internal representations are adapted to the task, the precision of $C_h$ is greatly improved, leading to an overall high precision extraction which converges at $0.96$--$0.99$. This development of extracting parallel pairs with increasing precision is in fact an instantiation of a \textbf{denoising curriculum} as described by \citet{wang2018denoising}.

The \textbf{recall} of the model, being bounded by the performance of the weakest representation, is very low at the beginning of training (R$\in$ [0.03,0.04]) due to the lack of task knowledge in $C_h$. However, as training progresses and $C_h$ improves, the accumulated extraction recall of the model rises to high values of $0.95$--$0.98$. Interestingly, the epoch-wise recall is much lower than the accumulated, which provides evidence for the hypothesis that \ssnmt\ models extracts different \emph{relevant} samples at different points in training, such that it has identified most of the relevant samples at some point during training, but not at every epoch.

It should be stressed that the successful extraction of increasingly precise pairs in combination with high recall is the result of the dynamics of both internal representations $C_w$ and $C_h$. As $C_h$ is less informative at the beginning of training, $C_w$ guides the final decision at such early stages to ensure high precision; and as $C_w$ is high in recall throughout training, $C_h$ ensures a gentle growth in final recall by setting a good lower bound. The intersection of both ensures that errors committed by one can be caught by the other; effectively a mutual supervision between representations. The results in Figure \ref{f:control} show that the \ssnmt\ self-induced curriculum is able to identify parallel data in comparable data with high precision and recall.

\begin{table*}[t]
 \small
 \centering
 \begin{tabular}{l rrr rrr rrr}
 \toprule
          & \#Pairs$_{enfr}$  & \en2\fr  & \fr2\en & \#Pairs$_{ende}$       & \en2\de      & \de2\en & \#Pairs$_{enes}$      & \en2\es      & \es2\en \\ \midrule 
 NMT$_{init}$ & 2.14M    & 21.8$\pm$.6   & 21.1$\pm$.5  &  0.32M   &  3.4$\pm$.3  & 4.7$\pm$.3 & 2.51M   & 27.0$\pm$.7  & 25.0$\pm$.7  \\
 NMT$_{mid}$  & 3.14M    & 29.0$\pm$.6  & 26.6$\pm$.6 &  1.13M    & 11.2$\pm$.4  & 15.0$\pm$.6  & 3.96M  & 28.3$\pm$.7 &  26.1$\pm$.7 \\
 NMT$_{end}$  & 3.17M   & 28.8$\pm$.6  & 26.5$\pm$.6 &  1.18M    & 11.9$\pm$.5   & 15.3$\pm$.5  & 3.99M & 28.3$\pm$.7 &  26.2$\pm$.7    \\
 NMT$_{all}$  & 5.38M    & 26.8$\pm$.7  & 25.2$\pm$.6 &  2.21M    &  11.6$\pm$.5  & 15.0$\pm$.6 & 5.41M & 27.9$\pm$.6 & 25.9$\pm$.8 \\
 \ssnmt\   & 5.38M    & 29.5$\pm$.6  & 27.7$\pm$.6 &  2.21M &  14.4$\pm$.6  & 18.1$\pm$.6 & 5.41M & 28.6$\pm$.7  & 28.4$\pm$.7 \\ 
  WikiMatrix & 2.76M & 33.5$\pm$.6 & 30.1$\pm$.6 & 1.57M & 13.2$\pm$.5 & 12.2$\pm$.5 & 3.38M & 29.6$\pm$.7 & 26.9$\pm$.8 \\ 
\bottomrule        
 \end{tabular}
 \caption{BLEU scores of a supervised NMT system trained on the unique pairs collected by \ssnmt\ in the first (NMT$_{init}$), intermediate (NMT$_{mid}$), final (NMT$_{end}$) and all (NMT$_{all}$) epochs of training tested on N13/N14.
 }
 \label{t:tq}
 \end{table*}
 
\paragraph{Comparison with WikiMatrix}

Because of the close similarity with our WP data, we compare on the $en$--$\{fr, de, es\}$ corpora in WikiMatrix \citep{schwenk2019wikimatrix}, which we pre-process as described in Section~\ref{s:data}. As these data sets consist of preselected mined sentence pairs together with their similarity scores, a manual threshold $\theta$ needs to be set to extract sentence pairs for training supervised NMT. We run the extraction script using $\theta=1.04$, which \citet{schwenk2019wikimatrix} recommend as a \emph{good choice for most language pairs}%
, and use the resulting data to train a supervised NMT system. 

The results are summarized in the bottom two rows in Table \ref{t:tq}. Confidence intervals ($p=95\%$) are calculated using bootstrap resampling \citep{koehn-2004-statistical}. For \en--\fr, the supervised system trained on WikiMatrix outperforms \ssnmt\ trained on WP by 3--4 BLEU points, while the opposite is the case for \en--\de, where \ssnmt\ achieves 1--5 BLEU points more. For \en--\es, both approaches are not statistically significantly different. The variable performance of the two approaches may be due to the varying appropriateness of the extraction threshold $\theta$ in WikiMatrix. For each language and corpus, a new optimal threshold needs to be found; a problem that \ssnmt\ avoids by its use of two representation types that complement each other during extraction without the need of a manually set threshold. The results
show that \ssnmt's self-induced extraction and training curriculum is able to deliver translation quality on a par with supervised NMT trained on externally preselected mined parallel data (WikiMatrix).

\begin{figure}[t]
  \includegraphics[width=1\linewidth]{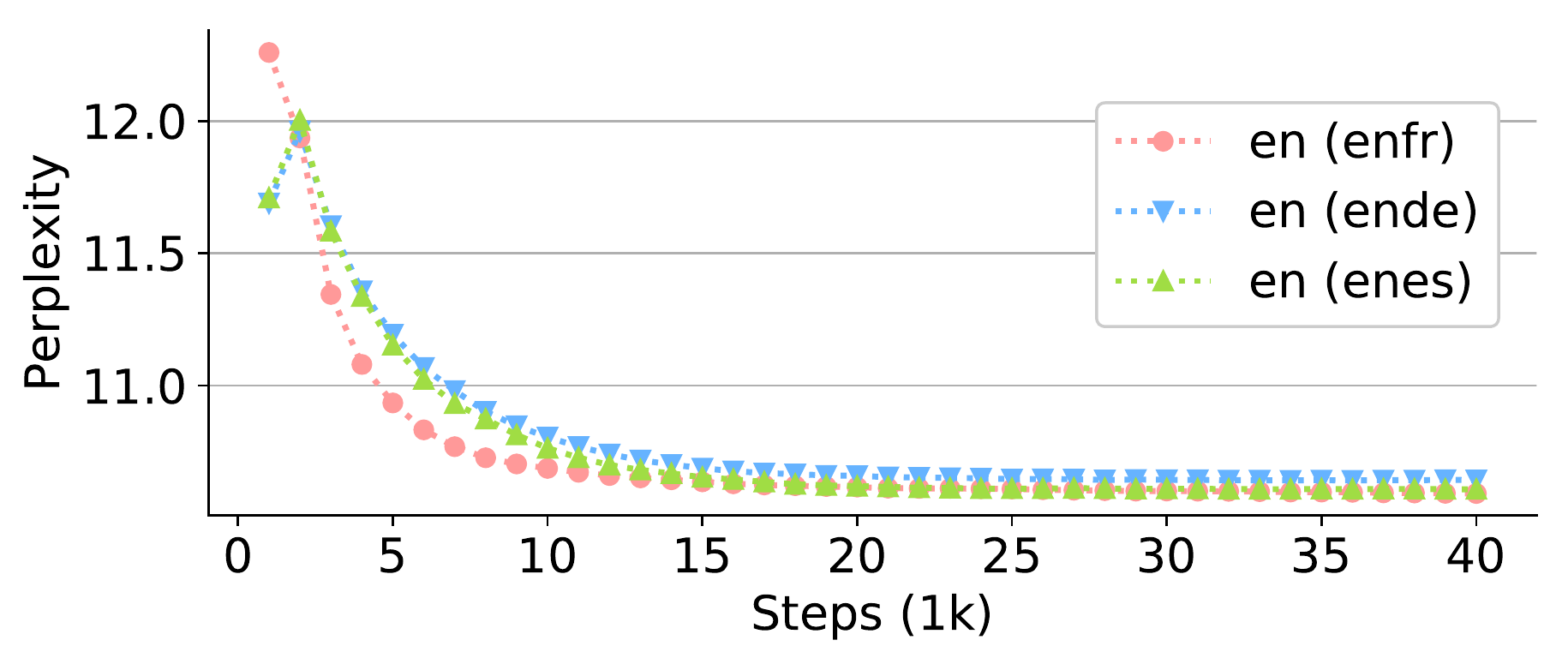}
  \hspace{-1.1em}
  \includegraphics[width=1\linewidth]{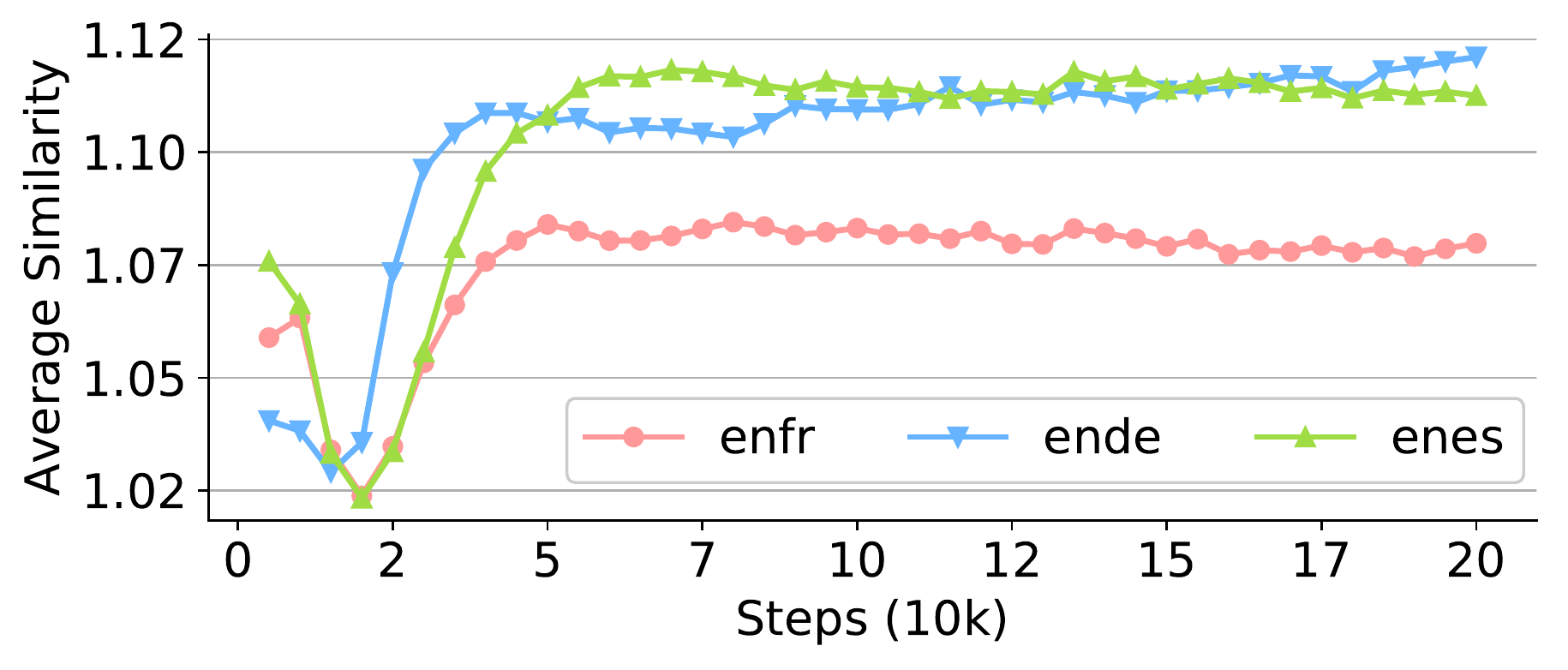}
\caption{
Perplexities on the English data extracted by \ssnmt\ (top) and average similarity scores of the accepted pairs (bottom).
}
\label{f:perp}
\end{figure}

\section{Self-Induced \ssnmt\ Curricula}
\label{s:cl}

\subsection{Order \& Closeness to the MT Task}
\label{ss:sim}

As a first indicator of the existence of a preferred choice in the order of the extracted sentence pairs, we compare the performance of \ssnmt\ with different supervised NMT models trained on the WP data extracted by \ssnmt\ at different points in training. We consider specific per-epoch data sets extracted in the first, intermediate and final epochs of training, as well as cumulative data of all unique sentence pairs extracted over all epochs. We then train four supervised NMT systems (NMT$_{init}$, NMT$_{mid}$, NMT$_{end}$, NMT$_{all}$) on these data sets. 
The difference in the {\bf translation quality} using only the data selected at different epochs reflects the evolving closeness of the data to the final translation task: we expect data extracted in later epochs of the \ssnmt\ training to include more sentences which are parallel, as demanded by a translation task, and therefore to achieve a higher translation quality.

For each language pair and system, the first four rows in Table~\ref{t:tq} show the number of sentence pairs extracted for training and the BLEU score achieved.
The evolving \ssnmt\ training curriculum outperforms all supervised versions across all tested languages. Notably, performance is 1--3 BLEU points above the supervised system trained on all extracted data, despite the fact that the \ssnmt\ system is able to extract only a small amount of data in its first epochs, compared to the fully supervised NMT$_{all}$, that, at every epoch, has access to all data that was ever extracted at any of the \ssnmt\ epochs. This suggests that the \ssnmt\ system is able to exclude previously accepted false positives in later epochs, while training supervised NMT on the complete data extracted by \ssnmt\ leads to a recurring visitation at each epoch of the same erroneous samples. 
Similar to a \textbf{denoising curriculum}, the quality and quantity of the extracted data grows as training continues for all languages, as the concatenation of the data extracted across epochs (NMT$_{all}$) is always outperformed by the last and thus largest epoch (NMT$_{end}$), despite the data for NMT$_{all}$\ being much larger in size.

An indicator of the \textbf{closeness of the curriculum to the final task} is the \textbf{similarity} between the selected sentence pairs during training.
We estimate similarity between pairs by their margin-based scores \citep{artetxe2018margin} during training.
At the beginning of training, the average similarity between extracted pairs is low, but it quickly rises within the first 100\,$k$ training steps to values close to $margin$ $1.07$ (\en--\fr) and $margin$ $1.12$ (\en--\{\de,\es\}). This evolution is depicted in Figure~\ref{f:perp} (bottom). The increase in mean similarity of the accepted pairs provides empirical evidence for our hypothesis that internal representations of translations grow closer in the cross-lingual space, and the system is able to exploit this by extracting increasingly similar and accurate pairs.

\begin{figure}[t]
  \includegraphics[width=0.97\linewidth]{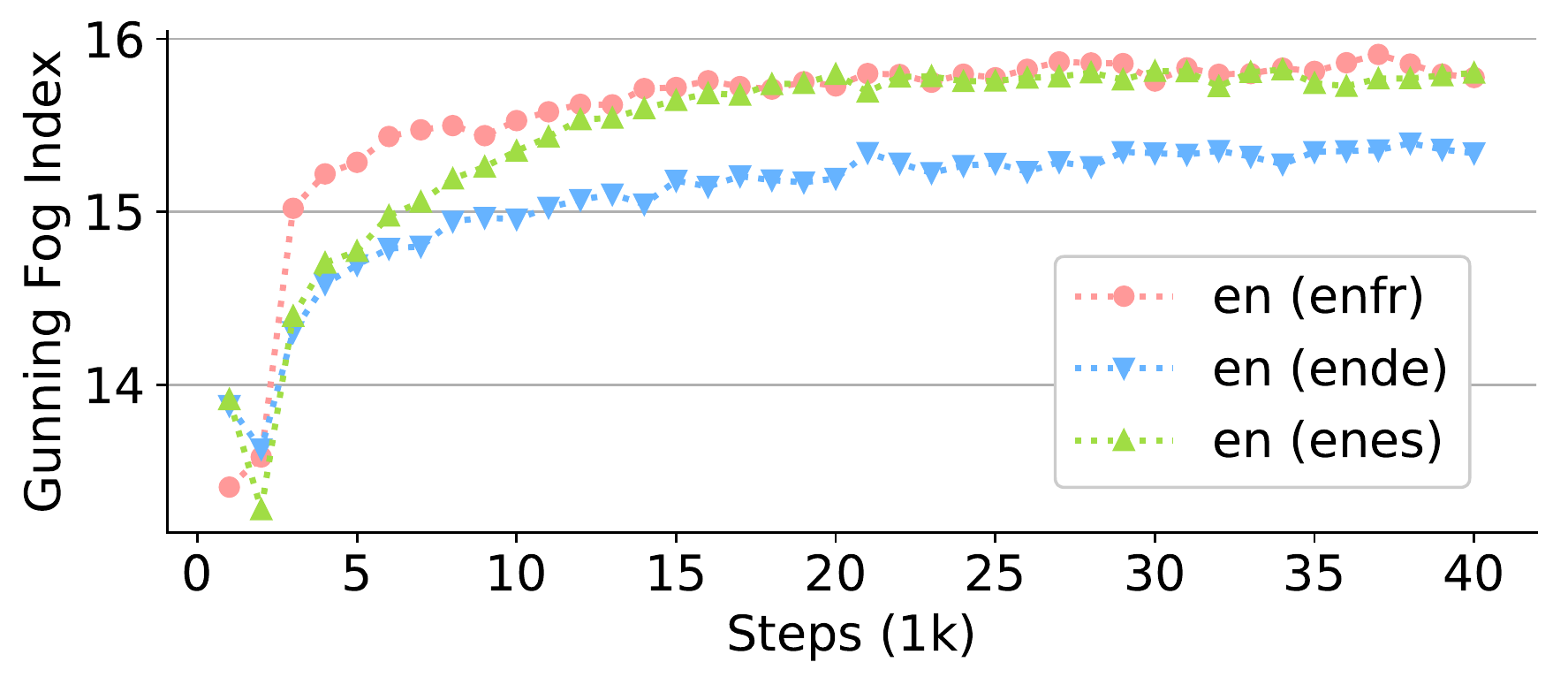} 
   \hspace{-1.2em}
  \includegraphics[width=0.98\linewidth]{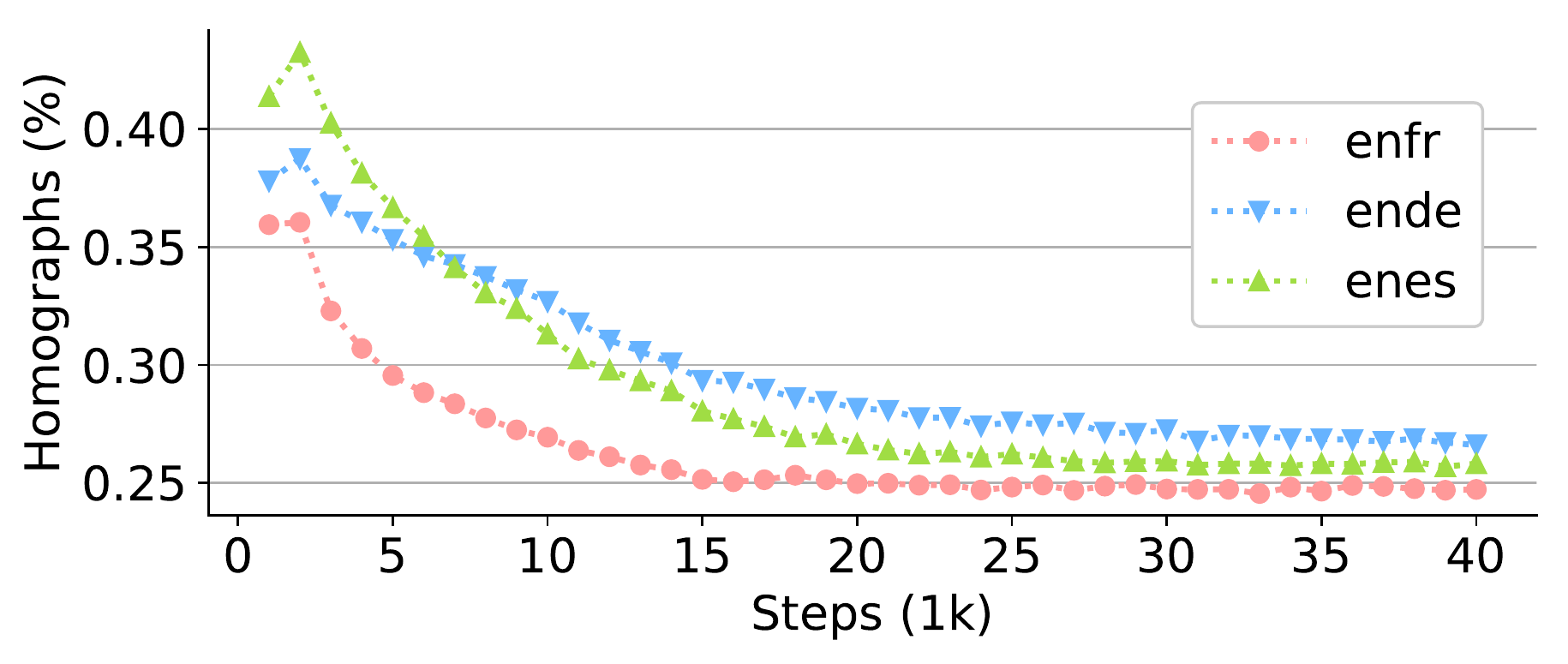} 
\caption{Gunning Fog Index (top) and percentage of homographs (bottom) of extracted English data seen during the first 40\,$k$ steps in training.}
\label{f:readability_homographs}
\end{figure}

\subsection{Order \& Complexity}
\label{ss:complexity}

Establishing the complexity of a sentence is a complex task by itself. Complexity can be estimated by the loss of an instance with respect to the gold or target. In our self-supervised approach, there is no target for the sentence extraction task, so we try to infer complexity by other means.

First, we study the behaviour of the average \textbf{perplexity} throughout training. Perplexities of the extracted data are estimated using a LM trained with \texttt{KenLM}
\citep{heafield:2011} on the monolingual WPs for the four languages in our study. We observe the same behaviour in the four cases illustrated by the English curves plotted in Figure~\ref{f:perp} (top). 
Perplexity drops heavily within the first 10\,$k$ steps for all languages and models. This indicates that the data extracted in the first epoch includes more \emph{outliers}, and the distribution of extracted sentences moves closer to the average observed in the monolingual WPs as training advances. The larger number of outliers at the beginning of training can be attributed to the larger number of homographs (bottom Figure \ref{f:readability_homographs}) and short sentences at the beginning of training, leading to a skewed distribution of selected sentences. 

The presence of \textbf{homographs} is vital for the self-supervised system in its initialization phase. At the beginning of training, only word embeddings, and therefore $C_w$, are initialized with pre-trained data, while $C_h$ is randomly initialized. Thus, words that have the same index in the shared vocabulary, homographs, play an important role in identifying similar sentences using $C_h$, making up around 1/3 of all tokens observed in the first epoch. As training progresses, and both $C_w$ and $C_h$ are adapted to the training data, the prevalence of homographs drops and the extraction is now less dependent on a shared vocabulary. The importance of homographs for the initialization raises questions on how \ssnmt\ performs on languages that do not share a script and it is left for future work.

\begin{figure*}
  \includegraphics[width=0.5\linewidth]{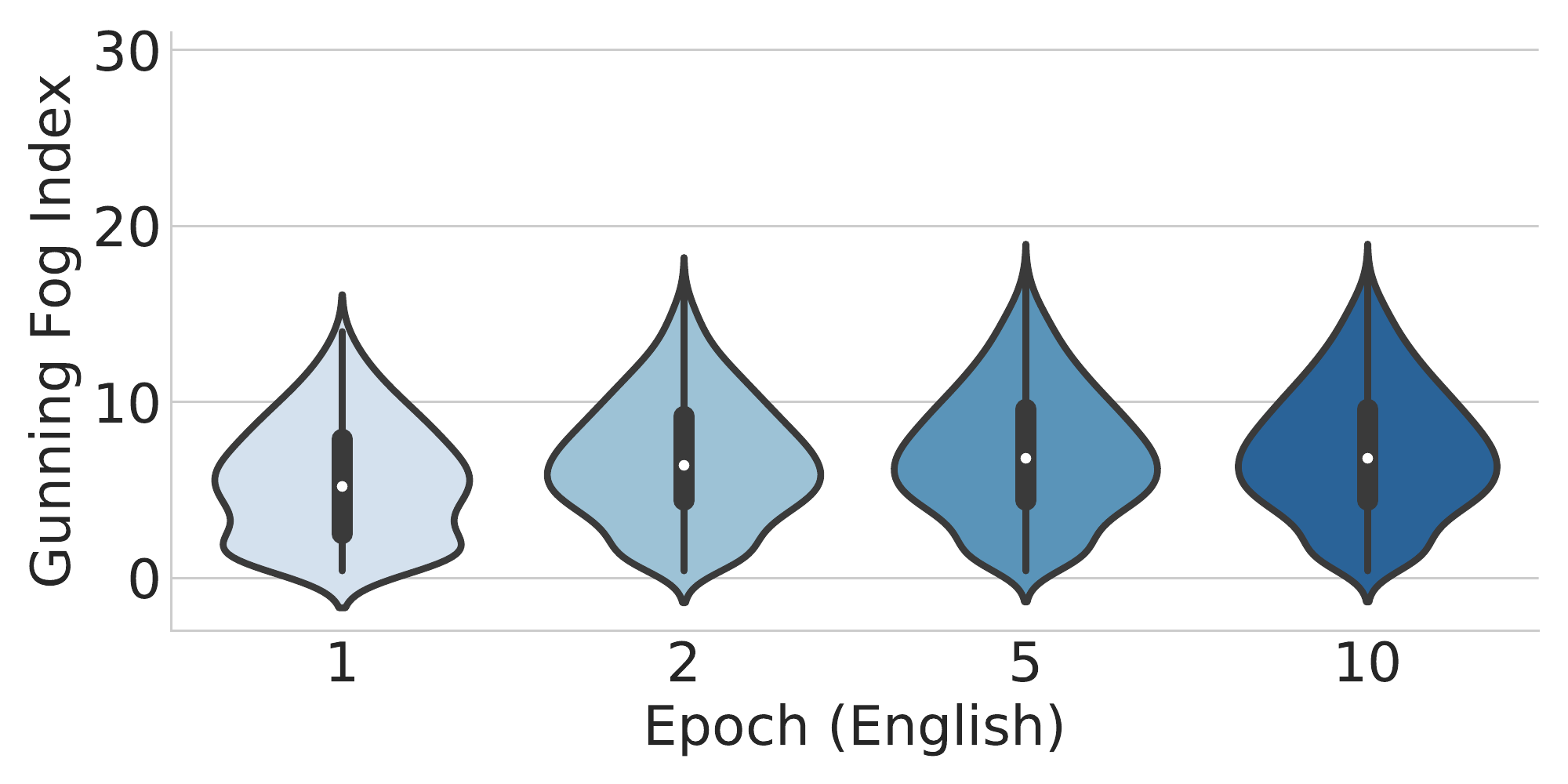}
\hspace{-0.5em}
  \includegraphics[width=0.5\linewidth]{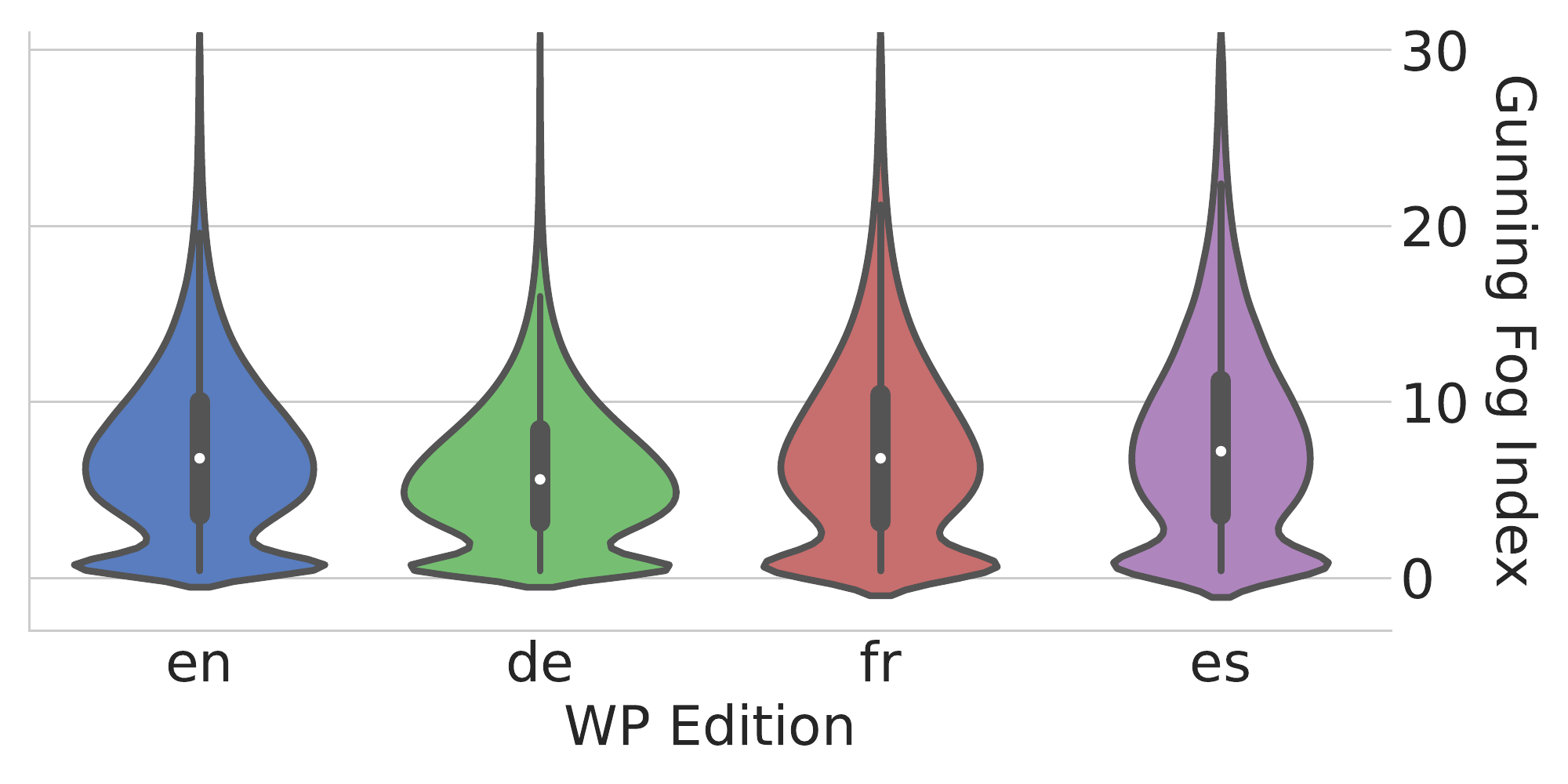}
\caption{Kernel density estimated Gunning Fog distributions and box plots over extracted \en\ (\en--\de) sentences at different points in training (left) and over the monolingual Wikipedias (right). 
}
\label{f:gf_dists}
\end{figure*}

Finally, we analyze the complexity of the sentences that an \ssnmt\ system selects at different points of training by measuring their \textbf{readability}. For this, we apply a modified version of the \textbf{Gunning Fog Index} (GF) \citep{gunning1952clear}, which is a measure predicting the years of schooling needed to understand a written text given the complexity of its sentences and vocabulary. It is defined as:
\begin{equation}
    \text{GF} = 0.4\left[\left(\frac{w}{s}\right) + 100\left(\frac{c}{w}\right)\right]
\end{equation}
where $w$ and $s$ are the number of words and sentences in a text. $c$ is the number of \emph{complex words}, which are defined as words containing more than $2$ syllables. The original formula excluded several linguistic phenomena from the complex word definition such as compound words, inflectional suffixes or familiar jargon; we do not apply all the language-dependent linguistic analysis.%

Since our training data is based on Wikipedia articles, the diversity in the complexity of the sentences is limited to the range of complexities observed in Wikipedia.
Figure~\ref{f:gf_dists}~(right) shows the per-sentence GF distributions over the sentences found in the monolingual WPs. We plot the probability density function for the sentence-level GF Index for the four WP editions estimated via a kernel density estimation. Each distribution is made up of two overlapping distributions: one at the lower end of the sentence complexity scale containing short article titles and headers, 
and one with a higher average complexity and larger standard deviation containing content sentences.

To study the behaviour during training, we compare the Gunning Fog distributions of the English data extracted at the beginning, middle and end of training \ssnmt$_{ende}$ with that of the original WP$_{en}$.
In the extracted data, we observe that compared with WP the overlapping distributions are less pronounced and that there is no trail of highly complex sentences. This is due to \Ni the pre-processing of the input data, which removes sentences containing less than 6 tokens, thus removing most WP titles and short sentences, and \Nii the length accepted in our batches, which is constrained to 50 tokens per sentence, removing highly complex strings. Apart from this, the distributions in the middle and the end of training come close to the underlying one, but we observe a large number of very simple sentences in the first epoch. This shows that the system extracts mostly simple content at the beginning of training, but soon moves towards complex sentences that were previously not yet identifiable as parallel.

A more detailed evolution is depicted in Figure \ref{f:readability_homographs} (top). We collect extracted sentences for each 1\,$k$ training steps and report their ``text''-level GF scores.\footnote{Note that GF is a text level score. In Figure \ref{f:gf_dists} we show sentence level GF distributions, while in Figure \ref{f:readability_homographs} (top) we show GF scores for ``texts'' consisting of sentences extracted over a 1\,k training step period.} Here we observe how the complexity of the sentences extracted rises strongly within the first 20\,$k$ steps of training. For English, most models start with text that is suitable for high school students (grade 10--11) and quickly turn to more complex sentences suited for first year undergraduate students ($\sim$13 years of schooling); a \textbf{curriculum of growing complexity}. The GF mean of the full set of sentences in the English Wikipedia is $\sim$12, which corresponds to a high school senior. For all other languages, a similar trend of growing sentence complexity is observed.

\subsection{Correlation Analysis}

\begin{figure}
    \centering
    \includegraphics[width=\linewidth]{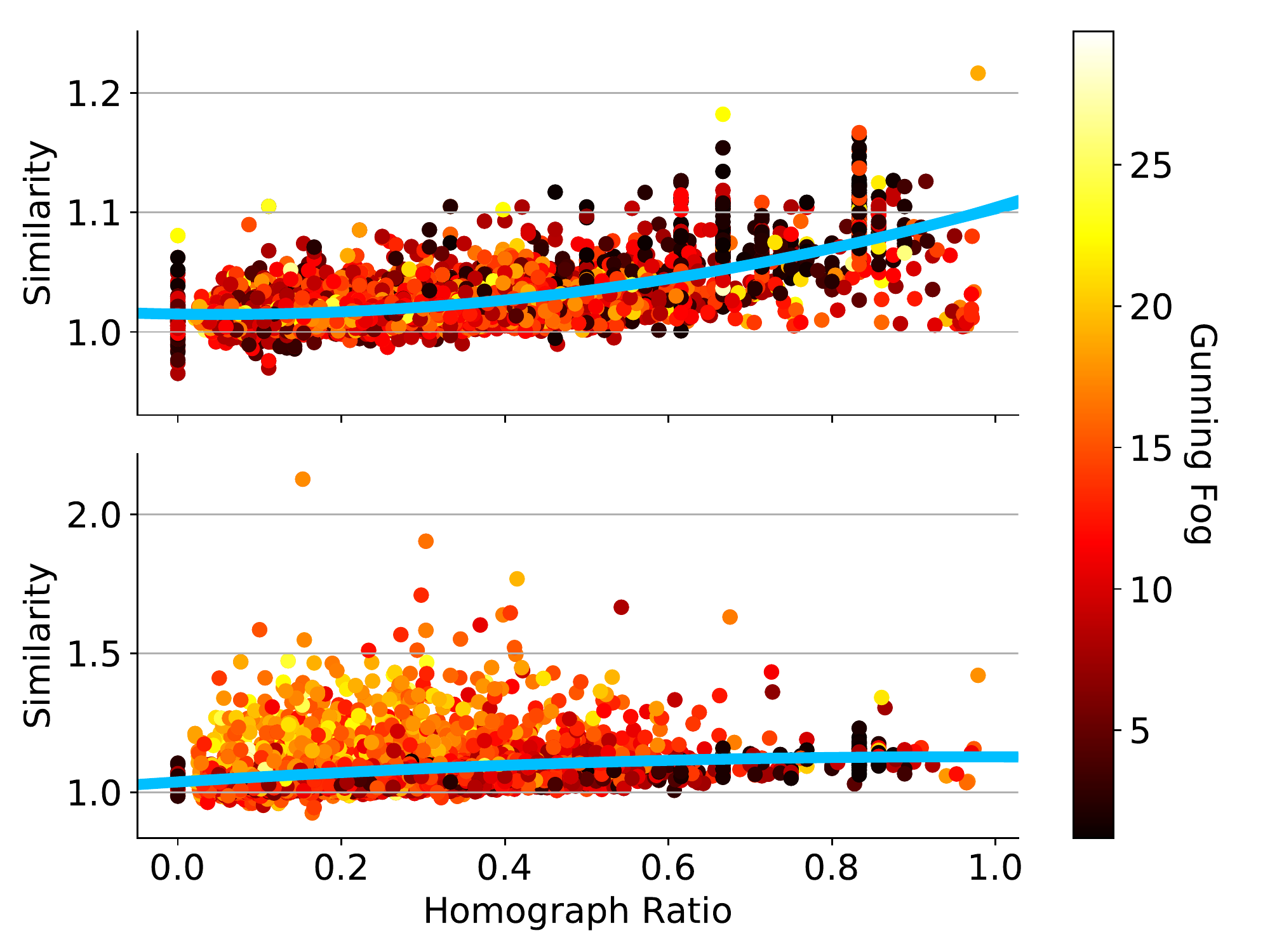}
    \caption{Margin-based similarity, homograph ratio and Gunning Fog index for the first 10\,$k$ extracted sentences in the first (top) and last (bottom) epoch of \en--\fr\ training. The solid blue line shows a second order polynomial regression  between the homograph ratio and similarity.}
    \label{fig:correlatons_enfr}
\end{figure}

So far, the variables under study,  similarity and complexity ---GF and homograph ratio---, have been observed as a function of the training steps.
In order to uncover the correlations between the variables themselves, we calculate the Pearson Correlation Coefficient ($r$) between them on the extracted pairs of the \en--\fr\ \ssnmt\ model during its first and last epoch. As shown in the previous sections of the paper, most differences appear in the first epoch and the behaviour across languages is comparable.

At the beginning of training (Figure \ref{fig:correlatons_enfr}, top) there is a positive correlation ($r=0.43$) between homograph ratio and similarity, naturally pointing to the importance of homographs for identifying similar pairs at the beginning of training. This is supported by a weak negative correlation between GF and homograph ratio ($r=-0.28$), indicating that sentences with more homographs tend to be less complex. While there is no significant correlation between GF and similarity in the first epoch ($r=-0.07$), in the last epoch of training (Figure \ref{fig:correlatons_enfr}, bottom), we observe a moderate positive relationship indicating that more complex sentences tend to come with a higher similarity ($r=0.30$). At this point, homographs become less important for the extraction and sentences without homographs are now also extracted in large numbers, indicated in terms of a weaker positive correlation between the homograph ratio and the similarity ($r=0.25$). The relationship between the homograph ratio and the GF stays stable ($r=-0.27$), as can be expected since the two values are not dependent on the MT model's state ($C_w$ and $C_h$), as opposed to the similarity score.

\section{Summary and Conclusions}
\label{s:conclusion}

This paper explores self-supervised NMT systems which jointly learn the MT model and how to find its supervision signal in comparable data; i.e. how to identify and select similar sentences. 
This association makes the system naturally and internally evolve its own curriculum without it having been externally enforced.
We observe that the dynamics of mutual-supervision of both system internal representations, $C_w$ and $C_h$, is imperative to the high recall and precision parallel data extraction of \ssnmt. Their combination for data selection over time instantiates a \textbf{denoising curriculum} in that the percentage of non-matching pairs, i.e. non-translations, decreases from 18\% to 2\%, with an especially fast descent at the beginning of training. 

Even if the quality of extraction increases over time, lower-similarity sentence pairs used at the beginning of training are still relevant for the development of the translation engine.
We analyze the translation quality of a supervised NMT system trained on the epoch-wise data extracted by \ssnmt\ and observe a continuous increase in BLEU.
Analogously, we also analyze the similarity scores of extracted sentences and observe that they also increase over time. 
As extracted pairs are increasingly similar, and precise, the extraction itself instantiates a secondary \textbf{curriculum of growing task-relevance}, where the task at hand is NMT learning with parallel sentences.

A tertiary \textbf{curriculum of increased sample complexity} is observed via an analysis of the extracted data's Gunning Fog indices. Here, the system starts with sentences suitable for initial high school students and quickly moves towards content suitable for first year undergraduate students: an overachiever indeed as the norm over the complete WP is end of high school level.

Lastly, by estimating the perplexity with an external LM trained on WP, we observe a steep decrease in perplexity at the beginning of training with fast convergence. This indicates that the extracted data quickly starts to resemble the underlying distribution of all WP data, with a larger amount of outliers at the beginning. These outliers can be accounted for by  the importance of homographs at that point. This raises the question of how \ssnmt\ will perform on really distant languages (less homographs) or when using smaller BPE sizes (more homographs), which is something that we will examine in our future work.

\section*{Acknowledgments}
The project on which this paper is based was funded by the German Federal Ministry of Education and Research under the funding code 01IW17001 (Deeplee). Responsibility for the content of this publication is with the authors.

\bibliography{anthology,emnlp2020}
\bibliographystyle{acl_natbib}

\end{document}